\pdfoutput=1
\PassOptionsToPackage{dvipsnames}{xcolor}
\PassOptionsToPackage{pdftex}{xcolor}
\documentclass[11pt]{article}

\usepackage{acl}

\usepackage{times}
\usepackage{latexsym}
\usepackage{color,soul}
\usepackage{booktabs}
\usepackage{multirow}
\usepackage{amsmath}  % for 'align' environment and other math features
\usepackage{amsfonts} % for '\mathbb' and other font-related commands
\usepackage{amssymb}  % for extra symbols
\usepackage{float}
\usepackage{inconsolata}
\usepackage{microtype}
\usepackage{algorithm}
\usepackage{algpseudocode}
\usepackage{smartdiagram}
\usepackage{appendix}
\usepackage{tikz}
\usepackage{tcolorbox}
\usetikzlibrary{arrows.meta, positioning, calc}
\usepackage{hwemoji}
\usepackage{hyperref}
\usepackage{xcolor}
\usepackage{subcaption}
\usepackage{amsthm}
\usepackage{bm}
\usepackage{pgfplots}
\pgfplotsset{compat=1.17}
\definecolor{olivegreen}{RGB}{107,142,35}
\definecolor{lightolivegreen}{RGB}{157,192,105}
\usepackage{graphicx}

\title{
\raisebox{-2.1ex}{\protect\includegraphics[height=4.5\fontcharht\font`\B]{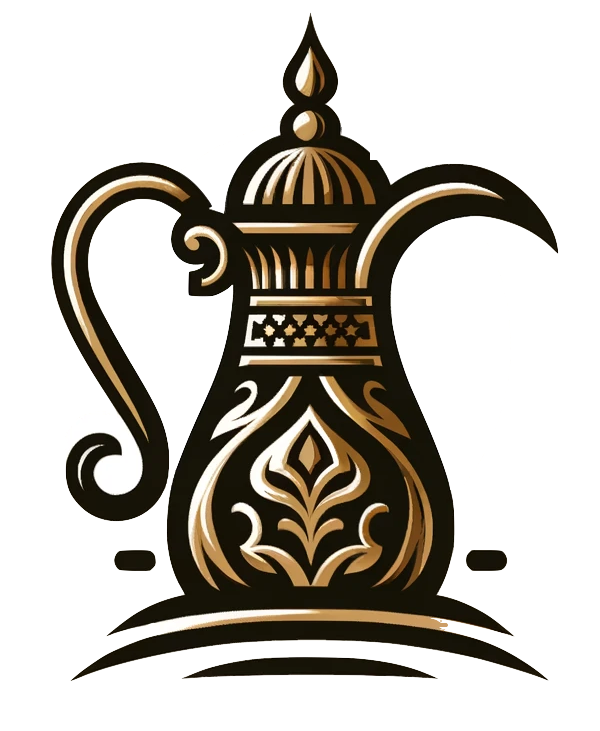}}
\textbf{Dallah}: A Dialect-Aware Multimodal Large Language Model for Arabic
}

\author{
\textbf{Fakhraddin Alwajih}~~~~  \textbf{Gagan Bhatia}~~~~\\ \textbf{Muhammad Abdul-Mageed}~~~~~
\\The University of British Columbia \& Invertible AI \\
\texttt{\normalsize \{fakhr.alwajih, muhammad.mageed\}@ubc.ca} \\  
}

\begin{document}
\maketitle
\begin{abstract}
Recent advancements have significantly enhanced the capabilities of Multimodal Large Language Models (MLLMs) in generating and understanding image-to-text content. Despite these successes, progress is predominantly limited to English due to the scarcity of high-quality multimodal resources in other languages. This limitation impedes the development of competitive models in languages such as Arabic. To alleviate this situation, we introduce an efficient Arabic multimodal assistant, dubbed \textit{Dallah}, that utilizes an advanced language model based on LLaMA-2 to facilitate multimodal interactions. \textit{Dallah} demonstrates state-of-the-art performance in Arabic MLLMs. Through fine-tuning six Arabic dialects, \textit{Dallah} showcases its capability to handle complex dialectal interactions incorporating both textual and visual elements. The model excels in two benchmark tests: one evaluating its performance on Modern Standard Arabic (MSA) and another specifically designed to assess dialectal responses. Beyond its robust performance in multimodal interaction tasks, \textit{Dallah} has the potential to pave the way for further development of dialect-aware Arabic MLLMs.
\end{abstract}
\section{Introduction}
% Contributions
% \begin{itemize}
%     \item Training Dallah based on LLaVA and using AraLlama as LLM with Arabic and English data.
%     \item Data filtering method
%     \item Support +5 Dialects by finetuning Dallah on limited dialectal data.
%     \item Human Evaluation Benchmark for dialects. 
% \end{itemize}

Large language models (LLMs) have revolutionized how machines understand and generate human language. Recent developments have significantly expanded the scope of these models by integrating multimodal data, enabling sophisticated interaction with both textual and visual information. Despite these advances, applying NLP in linguistically diverse environments presents unique challenges, particularly in processing dialectal variations of languages and the integration of these variations within multimodal contexts. These challenges are especially pronounced in Arabic, a collection of languages and varieties characterized by a rich tapestry of dialects that vary significantly across different regions.

% \mam{Can we stop depending on ChatGPT for writing? It is very verbose and uses certain phrases repeatedly. I have removed "In the dynamic field of Natural Language Processing (NLP), the advent of large language models (LLMs)" and just said "Large language models (LLMs)" directly.}
Arabic dialects enrich the cultural landscape and present complex linguistic variations that standard NLP models, primarily designed for MSA, often fail to solve. This linguistic diversity requires the development of specialized models that can navigate the rich world of dialectal Arabic and its integration with visual data. Addressing these needs is crucial for improving user interaction and preserving linguistic heritage, as many dialects are underrepresented. Some may be at risk of diminishing in the face of globalization and cultural homogenization.
% \mam{I like this last statement. Can we find a couple of references about this concept of "cultural homogenization"?}

Multi-cultural and multi-modal LLMs
\citep{alwajih2024peacock, huang2023acegpt, sengupta2023jais} are vital against cultural homogenization, where globalization tends to favour dominant languages like Arabic. This  can potentially lead to the marginalization or even extinction of less widely spoken dialects~\cite{barnet2014homogenization, ahmedov2024english, fahmi2024promoting}. By including these low-resource dialects, we can ensure their continued relevance and preserve the rich linguistic diversity of the Arabic-speaking world.

\begin{figure}[t]
    \centering
    \includegraphics[width=\linewidth]{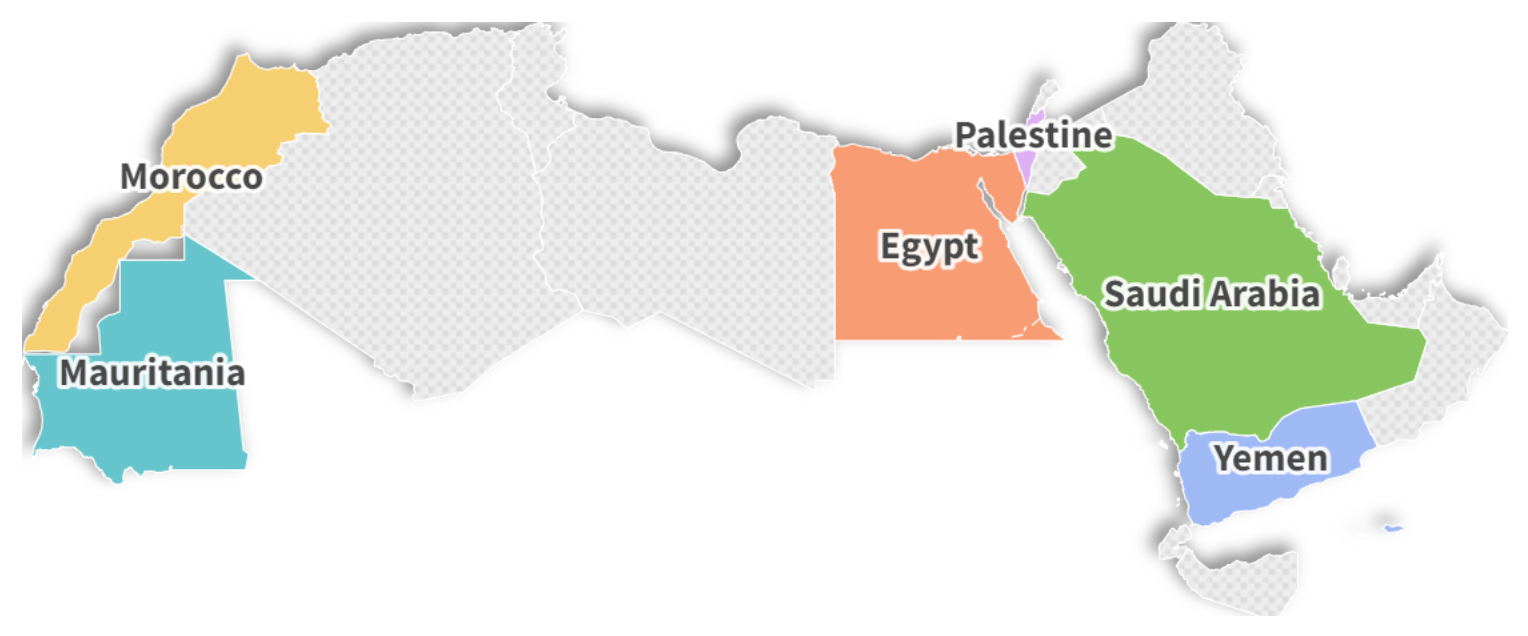}
    \caption{Map highlighting the countries targeted by \textit{\textit{Dallah}} for dialectal Arabic dataset construction.}
    \label{fig:countries}
\end{figure}

To this end, we introduce a powerful multimodal language model that specifically targets the unique aspects of Arabic dialects. Our model, dubbed~\textit{Dallah}, is built on the foundations of LLaVA~\cite{liu2023llava}, an advanced multimodal language model framework. We enhance LLaVA with the linguistic capabilities of AraLLaMA~\cite{alwajih2024peacock}, an LLM proficient in Arabic and English. \textit{\textit{Dallah}} is designed to understand and generate content in Arabic, navigating the complex interplay between different dialects and visual information effectively.
% By providing robust support for low-resource dialectal Arabic varieties, \textit{\textit{Dallah}} helps preserve linguistic heritage, ensuring that the rich tapestry of Arabic dialects is maintained for future generations. 

The following are the main contributions of our work:
\begin{enumerate}
    \item We present \textit{\textit{Dallah}}, which combines the robust multimodal processing power of LLaVA with the dialectal versatility of AraLLaMA, creating a model uniquely equipped to handle the linguistic and visual challenges presented by Arabic dialects.
    \item We introduce a novel data filtering method that optimizes the selection and usage of training data, ensuring that \textit{\textit{Dallah}} is fine-tuned with high-quality, relevant multimodal datasets that reflect the linguistic diversity found within the Arab world.
    \item \textit{Dallah} supports wide dialectal coverage, successfully fine-tuning over six major Arabic dialects using limited but highly representative dialectal data. 
    \item We introduce \textit{Dallah-Bench} evaluation benchmark for Arabic dialects tailored to assess the efficacy of multimodal language models in real-world applications that require an understanding of dialectal variations.
    \item We have also built an understanding of which model from the set \textit{\{GPT4, GPT4-Turbo, Command-R+\}} is best suited for evaluating MSA and dialectal data compared to Human evaluation. 
\end{enumerate}

The remainder of this paper is structured as follows: In Section~\ref{sec:related_work}, we provide an overview of related work. Section~\ref{sec:methodology} details our methodology, including the processes for Arabic dataset translation and filtering, construction of dialectal Arabic datasets, the architecture of \textit{\textit{Dallah}}, and the training procedures employed. In Section ~\ref{sec:experiments}, we describe our implementation details and the benchmarks used for evaluation. Section~\ref{sec:results} presents our experimental results, including both quantitative and qualitative analyses. We conclude in Section~\ref{sec:conclusion} with a discussion of our findings and future work.

% This comprehensive approach advances the frontiers of Arabic NLP. It contributes significantly to the broader field of language technology by showcasing how multimodal systems can support and enrich the preservation and understanding of linguistic diversity.

\section{Related Work}
\label{sec:related_work}
\subsection{Large Language Models}
Recent progress in NLP has been driven by advances in LLMs, starting with the foundational Transformer model~\cite{vaswani2017attention}. This innovation paved the way for language models like the encoder-based BERT~\cite{devlin2018bert}, and the decoder-based Generative Pre-trained Transformer (GPT)~\cite{brown2020language}, as well as encoder-decoder-based models like T5~\cite{raffel2020exploring}, which have significantly improved linguistic understanding and performance in complex language processing tasks. The development of models such as OPT~\cite{zhang2022opt}, LLaMA~\cite{touvron2023llama}, LLaMA-2~\cite{touvron2023llama2}, GPT-2~\cite{radford2019language}, GPT-3~\cite{brown2020language}, GPT-4~\cite{achiam2023gpt}, and ChatGPT~\cite{openai2023chatgpt} Mistral~\cite{jiang2023mistral}, Mixtral~\cite{jiang2024mixtral}, Phi-2~\cite{javaheripi2023phi2}, Phi-3~\cite{abdin2024phi}, and instruction-tuned variants of LLaMA-2 like Alpaca~\cite{alpaca} and Vicuna~\cite{vicuna2023}, have demonstrated the rapid evolution of the field. These models benefit from extensive training on large datasets and tailored instruction sets, enhancing their effectiveness.

\paragraph{Arabic LLMs}
Building on the global momentum, the scope of LLMs has extended into Arabic language processing. The introduction of Jasmine \cite{nagoudi2023jasmine} marked a significant milestone, followed by AceGPT \cite{huang2023acegpt} and Jais \cite{sengupta2023jais}, which have enhanced Arabic conversational AI. Recently, AraLLaMA \cite{alwajih2024peacock} set a new standard with its proficiency in the Egyptian Arabic dialect, showcasing the flexibility of LLMs in handling linguistically diverse data.

\subsection{Multimodal Large Language Models}
The integration of computer vision and natural language processing has given rise to Visual Language Models (VLMs). These models merge visual and linguistic data, enhancing tasks that require visual perception and language abilities. Models like CLIP~\cite{radford2021learning} bridge the gap between visual recognition and language tasks, demonstrating the effectiveness of cross-modal applications.

Recent advancements show that LLMs improve VLMs. Innovations such as Flamingo~\cite{alayrac2022flamingo}, Blip-2~\cite{li2023blip}, and LLaVA~\cite{liu2023llava} have leveraged large image-text pair datasets, enhancing cross-modal coordination and learning efficiency. These models also employ specialized architectural features for better integration. For instance, Flamingo utilizes a perceiver resampler to integrate visual data, and Blip-2 introduces a Q-Former~\cite{li2023blip} for aligning visual and language modalities. LLaVA~\cite{liu2023llava} adjusts a linear projection layer to synchronize vision and language modalities. Meanwhile, LLaVA1.6~\cite{liu2023improvedllava} incorporates extensive instruction tuning and a high-resolution vision encoder, achieving outstanding results across multiple benchmarks. 
% The intersection of visual data processing and language understanding has given rise to Visual Language Models (VLMs), such as CLIP \cite{radford2021learning}, which bridge the modalities of vision and text. Recent innovations have seen LLMs enhancing VLMs, as demonstrated by Flamingo \cite{alayrac2022flamingo}, Blip-2 \cite{li2023blip}, and LLaVA \cite{liu2023llava}, which leverage extensive image-text pair datasets for improved cross-modal coordination.

\paragraph{Arabic Multimodal LLMs}
In Arabic NLP, Peacock \cite{alwajih2024peacock} represents the first work in Arabic-centric MLLM capable of handling  Arabic multimodal interaction effectively. Additionally, the multilingual PALO \cite{maaz2024palo} has demonstrated the ability to process and integrate multiple languages, including Arabic, into multimodal contexts.

\subsection{Multimodal Instruction Tuning Datasets}
Development of MLLMs typically involves two phases. The first phase focuses on aligning visual and linguistic features, utilizing datasets such as COCO~\cite{lin2014microsoft}, LLaVA-Pretrain~\cite{liu2023llava}, and Laion \cite{schuhmann2022laion}. The subsequent visual instruction fine-tuning phase enhances the models' capabilities to follow complex multimodal instructions. This phase often involves transforming existing datasets into more conversationally relevant formats using advanced LLMs such as GPT-4, as seen in models like LLaVA-Instruct~\cite{liu2023llava} and SVIT~\cite{liu2023improved}. Recent works utilized GPT-4V~\cite{openai2023chatgpt} to generate new captions and question-answers, such as in ShareGPT4V~\cite{chen2023sharegpt4v}, LVIS-instruct4v~\cite{wang2023see} and Allava~\cite{chen2024allava}. Arabic MLLMs utilized translated versions of LLaVA-Instruct~\cite{alwajih2024peacock,maaz2024palo} using different tools for translation. 

% Our work with "Dallah" integrates these advancements, presenting a unique framework that leverages the specificity of Arabic dialects in a multimodal context, pushing the boundaries of what is possible with language and vision models in Arabic NLP.

\section{Methodology}
\label{sec:methodology}

\subsection{Arabic Dataset Translation and Filtering}
\label{sec:translate_filter}

\begin{figure*}[!htb]
    \centering
    \includegraphics[width=0.9\linewidth]{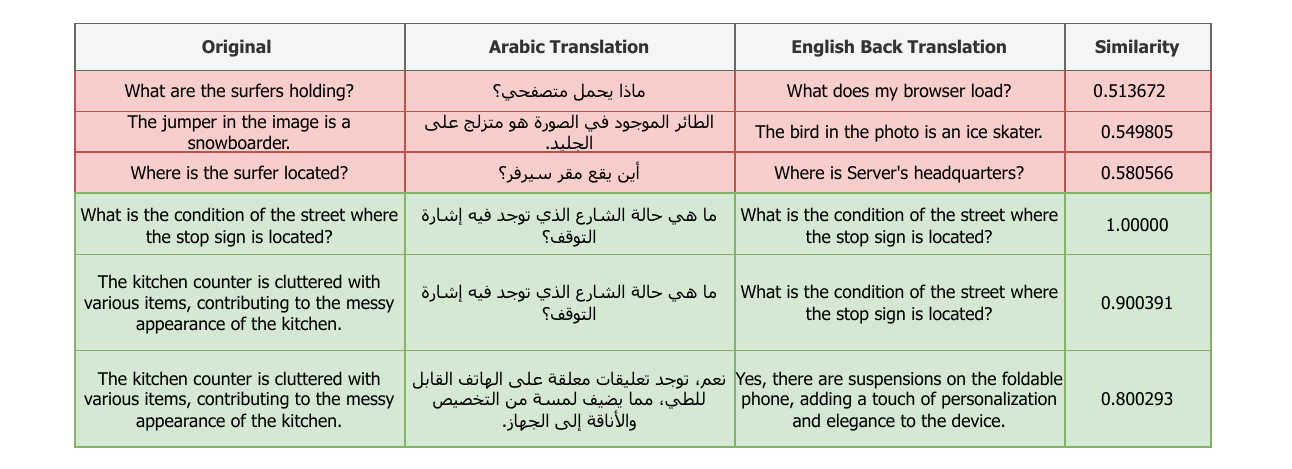}
    \caption{This figure illustrates the translation and filtering process used in constructing the Arabic dataset for \textit{Dallah}. The red rows represent examples that were removed due to low similarity scores between the original English text and the back-translated English text. The green rows show the retained examples that met the similarity threshold, ensuring high-quality translations for effective model training.}
    \label{fig:trans_translate_filter}
\end{figure*}
In the first step, we aimed to build an Arabic MLLM using Arabic datasets. A major obstacle facing Arabic MLLMs is the lack of resources. This lack is largely due to the challenges of sourcing relevant Arabic image-text pairs on a large scale. To bridge this resource gap, we have implemented a careful translate-and-filter pipeline consisting of a translation stage and a filtering stage inspired by~\cite{mohamed2023violet,alwajih2024peacock}. This pipeline converts publicly available, English-centric image-text and visual instruction datasets into Arabic while maintaining data quality and preventing error propagation due to translation.

\begin{figure}[!htb]
    \centering    \includegraphics[width=0.9\linewidth]{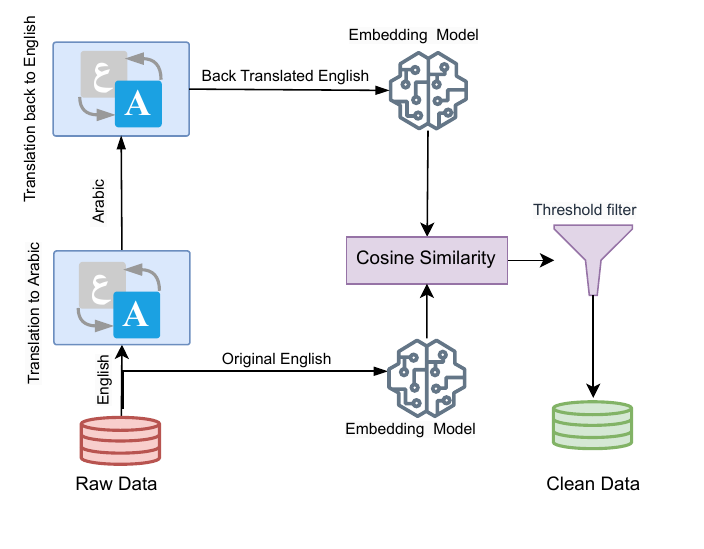}
    \caption{Illustration of the translation and filtering process for constructing high-quality Arabic multimodal datasets.  Examples illustrating the results of this pipeline are in Figure \ref{fig:trans_translate_filter}.}
    \label{fig:translate_filter}
\end{figure}

We utilize the latest version of the Google Translate API (Google Cloud) for \textbf{the translation stage} which is the best translation method as shown by \cite{zhu2023multilingual}. We also conducted back translation as required by the subsequent filtering stage. During \textbf{the filtering stage}, we ensure the quality of our translations by employing a sentence embedding model~\cite{SFRAIResearch2024,wang2024multilingual}. We assess the quality by calculating the similarity of embedding between the original and back-translated sentences for both question and answer pairs, retaining only those translations that meet our quality standards. Essentially, we keep examples with questions and answers above a predefined threshold, which we have empirically set at 80\%. Figure \ref{fig:translate_filter} illustrates the translation and filtering process. Unlike the methods used in ~\cite{mohamed2023violet,alwajih2024peacock}, we employ an English sentence embedding model based on Mistral-7b~\cite{wang2024multilingual} to calculate the similarities. This last model is more powerful than embedding models used in ~\citet{mohamed2023violet,alwajih2024peacock} as shown by MTEB leaderboard \citep{muennighoff2022mteb}. Refer to \autoref{fig:trans_translate_filter} for examples illustrating our pipeline's effectiveness. 

\subsection{Dialectal Arabic Dataset Construction}
Due to the absence of Arabic dialectal data tailored for vision tasks, we randomly select six subsets from our translated LLaVA-instruct 150k dataset. We aim to ensure that each subset included diverse content types, such as conversations, complex reasoning, and detailed descriptions, from the LLaVA-instruct dataset. Our goal is to capture the dialects of Egypt, Mauritania, Morocco, Palestine, Saudi Arabia, and Yemen. These dialects represent a broad spectrum of Arabic dialects.

These subsets are then assigned to native professional translators from the aforementioned countries in order to translate them from MSA into their respective dialects. Table~\ref{tab:dialects_dataset} displays the number of samples per country, while Figure~\ref{fig:countries} illustrates the targeted countries on the map. Refer to ~\ref{apdx:dialect_examples} for more details. 

\begin{table}[H]
    \centering
    \resizebox{\columnwidth}{!}{%
    \begin{tabular}{llr}\toprule
\textbf{Stage} &\textbf{Source} &\textbf{\#Sample} \\\midrule
Pretraining & LLaVA-Pretrain LCS (Arabic+English)&800k  \\\midrule
\multirow{2}{*}{Inst. Tuning } &LLaVA-Instruct English&150k \\
&LLaVA-Instruct  Arabic&139k \\\midrule
\multirow{6}{*}{Dialectal Tuning} &Egypt &738 \\
&Mauritania &495 \\
&Morocco &505 \\
&Palestine &853 \\
&Saudi Arabia &784 \\
&Yemen &604 \\\midrule
\multicolumn{2}{c}{Total} &1.1 M \\
\bottomrule
\end{tabular}
    }
    \caption{Number of samples used for each stage in the training of \textit{Dallah}.}
    \label{tab:dialects_dataset}
\end{table}
% \begin{figure}[H]
%     \centering
%     \includegraphics[width=0.9\linewidth]{figs/map.pdf}
%     \caption{Countries}
%     \label{fig:countries}
% \end{figure}

\subsection{Architecture}
The \textit{Dallah} model follows the structure of LLaVA1.5~\cite{liu2023improvedllava} and comprises three key elements:

\begin{enumerate}
    \item \textbf{Vision Encoder:} The vision encoder ($V_{\varphi}$) employs the CLIP-Large model~\cite{radford2021learning} to process input images ($X$) into 576 visual tokens at a resolution of 336x336 with a patch size of 14, producing a sequence of patch features $V = \{v_j \in \mathbb{R}^{d_x}\}_{i=j}^{M}$.
    \item \textbf{Projector:} A connector ($P_{\phi}$), designed as a two-layer multi-layer perceptron (MLP), maps the visual patch sequences $\{v_j\}_{j=1}^{M}$ to the text embedding space $\{h_j\}_{j=1}^{M}$, allowing for effective integration between the pre-trained LLM and the vision encoder.
    \item \textbf{Language Model (LLM):} The Arabic LLM ($F_{\theta}$), based on AraLLaMA~\cite{alwajih2024peacock}\footnote{AraLLaMA is an enhanced version of LLaMA-2~\cite{touvron2023llama}.}  processes sequences of text embeddings $\{h_i\}_{i=0}^{N-1}$ in the $d$-dimensional space, outputting corresponding next predictions $\{h_i\}_{i=1}^{N}$. A tokenizer and embedding module maps text sequences $\{y_i\}_{i=0}^{N-1}$ to the embedding space and back to output text sequences $\{y_i\}_{i=1}^{N}$.
\end{enumerate}

\begin{figure}[!htb]
    \centering
    \includegraphics[width=\linewidth]{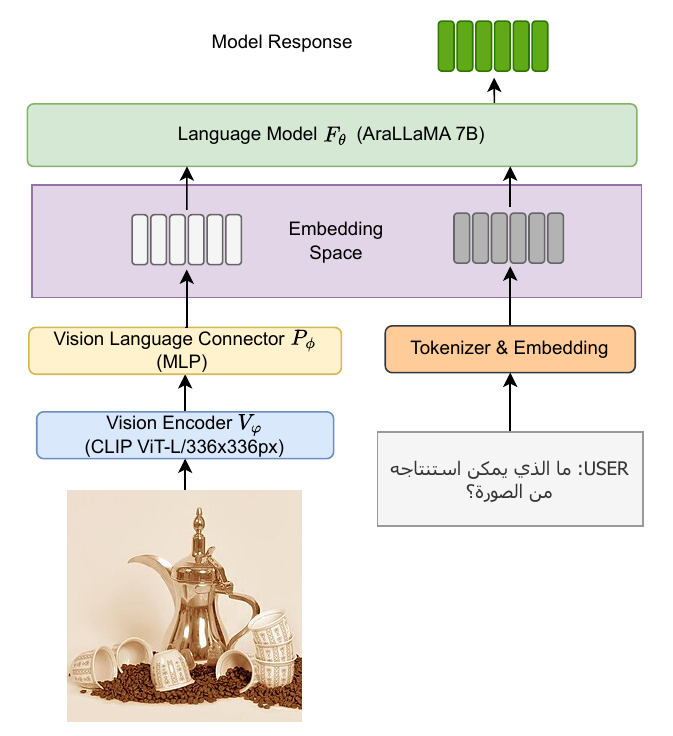}
    \caption{\textit{\textit{Dallah}} model architecture, showcasing the integration of the vision encoder, projector, and language model.}
    \label{fig:model_arch}
\end{figure}

This structure equips the model to handle various multimodal understanding tasks, taking an image and instruction text sequence as input and generating a text sequence as output. Figure~\ref{fig:model_arch} illustrates \textit{Dallah} architecture. 

% The Dallah model follows the structure of LLaVA1.5~\cite{liu2023improvedllava} and consists of three key components: (1) A vision encoder using the CLIP-Large model~\cite{radford2021learning}, which converts input images into 576 tokens at a resolution of 336x336 with a patch size of 14. (2) A projector, a two-layer multi-layer perceptron (MLP), connecting the vision and language modalities. (3) A language model (LLM) based on the AraLLAMA, derived from LLaMA2~\cite{touvron2023llama}. This design enables the model to handle various multimodal understanding tasks, taking an image and text sequence as input and producing a text sequence as output.
% , for which we expanded the vocabulary of the tokenizer and continued pre-training using a combination of Arabic and English data, then subsequently fine-tuned it. 

\subsection{Training}
Training of \textbf{Dallah} consists of three stages: (i) pre-training using data LLaVA-Pretrain (MAS Arabic and English), (ii) visual instruction supervised fine-tuning using LLaVA-Instruct (Arabic MSA and English), and (iii) further visual instruction supervised fine-tuning using dialectal data. Table \ref{tab:dialects_dataset} details the datasets used in each stage.
Training data comprises pairs of images and text \((X, Y)\), with the text sequence \(Y\) formatted as a single-turn in the pre-training phase and multi-turn conversation in the visual Instruction Supervised Fine-tuning stage.  
\(Y = (Y_q^1, Y_a^1, \ldots, Y_q^T, Y_a^T)\). Here, \(T\) represents the number of conversation turns, \(Y_q^t\) the user's prompt, and \(Y_a^t\) the model's response.

\begin{figure}[!htb]
    \centering
    \includegraphics[width=\linewidth]{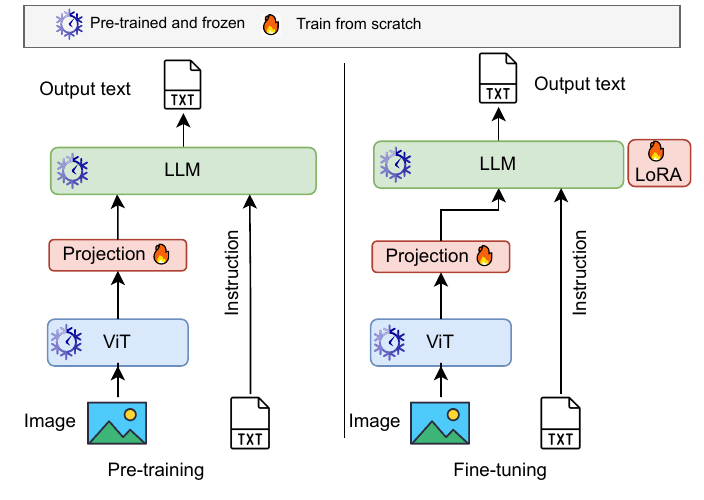}
    \caption{Training schema for \textit{\textit{Dallah}}, detailing the pre-training and visual instruction supervised fine-tuning phases.}
    \label{fig:training_schema}
\end{figure}

\subsubsection{Pre-training}
\label{pretraining}

During this phase, the goal is to enhance the alignment between the vision and text data within the embedding space. For this, image-caption style data \((X, Y_a)\) is extracted from the conversation, where \(X\) is the image and \(Y_a\) is a corresponding text description. The probability of generating \(Y_a\) given the image is calculated as:
\[
p(Y_a | X) = \prod_{i=1}^{N_a} F_{\theta}(y_{i} | P_{\phi} \circ V_{\varphi}(X)),
\]
where \(N_a\) is the length of \(Y_a\). The training objective is to maximize the log-likelihood of \(Y_a\) autoregressively:
\[
\max_{  \phi} \sum_{i=1}^{N_a} \log F_{\theta}(y_{i} | P_{\phi} \circ V_{\varphi}(X)),
\]
This framework permits the adjustment of learnable parameters of projector layers during pre-training. LLM and vision encoder learnable parameters are frozen during pre-training, as shown in Figure~\ref{fig:training_schema}.

\subsubsection{Visual Instruction Supervised Fine-tuning}
\label{finetuning}
The full image-text pairs \((X, Y)\) in their multi-turn conversation format are used for fine-tuning. The set of tokens corresponding to the model's responses is denoted as \(A = \{y | y \in Y^t_a, \text{for any } t = 1, \ldots, T\}\). The training objective is to maximize the log-likelihood of the model's responses autoregressively:
\[
\max_{\acute{\theta}, \phi} \sum_{i=1}^{N} I(y_i \in A) \log  F_{\theta}(y_{i} | P_{\phi} \circ V_{\varphi}(X)),
\]
where \(N\) is the total number of tokens in \(Y\),  \(\acute{\theta}\) is a subset of \(\theta\) , and \(I(y_i \in A)\) is \(1\) if \(y_i\) belongs to \(A\), and \(0\) otherwise.  Training the projector layers' learnable parameters while the vision encoder is kept frozen during this phase. For LLM, we utilize LoRA~\cite{hu2021lora} to train the LLM. Figure~\ref{fig:training_schema} visualizes both pre-training and visual instruction supervised fine-tuning stages.

\section{Experiments}
\label{sec:experiments}
\subsection{Implementation Details}
\paragraph{Model Configurations.} In this work, we develop the \textit{Dallah} model based on the LLaVA1.5 framework. Specifically, we employ the CLIP-ViT-L/14 as the visual encoder with a standard resolution of \(336 \times 336\). We also use AraLLaMA, a language model tailored specifically for Arabic, and employ a two-layer MLP as the interface to connect the visual encoder with the LLM.

For the construction of \textit{Dallah}, a three-stages training process is implemented. Initially, we establish a base MSA MLLM in two stages pretraining and MSA fine-tuning, followed by an adaptation stage for Arabic dialects. The following are details of these different stages:

\paragraph{Stage 1: Pre-training stage.} During this initial stage, training is conducted for a single epoch only on the projector as described in~\ref{pretraining}  using the translated LCS-558~\cite{liu2023llava} dataset. This dataset includes data filtered for Arabic and 300K samples of English samples. The optimization is done using the AdamW optimizer with a learning rate of \(1 \times 10^{-3}\), combined with a cosine learning rate schedule. The overall batch size is maintained at 32. This phase required approximately 28 hours of training on a single A100 GPU.

\paragraph{Stage 2: Instruction-tuning stage.} In this satge, the visual encoder is frozen, tuning is applied to the visual projector, and the LLM is fine-tuned using LoRA as described in \ref{finetuning}. Here we employ the 150K LLaVA-Instruct dataset in English alongside a translated and filtered Arabic version. The learning rate is set to \(2 \times 10^{-4}\) with a batch size of 8, maintaining the same settings as in the first stage. This training phase took around 58 hours using a single A100 GPU.

\paragraph{Stage 3: Dialectal instruction-tuning stage.} This stage is similar to stage 2, but is focused on dialectal data for six different Arabic dialects and parallel data for MSA. The settings remain the same as in the second stage with learning rate \(2 \times 10^{-5}\), over five epochs. This training phase took approximately 2 hours using a single A100 GPU. Table~\ref{tab:dialects_dataset} details the data used in the aforementioned stages.

\subsection{Benchmarks}
We evaluate our model using two benchmarks: LLaVA-Bench for MSA evaluation and comparison with counterpart Arabic MLLMs and \textit{Dallah-Bench} to assess the model's capabilities in six Arabic dialects.
\subsubsection{LLaVA-Bench for MSA}
\label{llava_bench}
The Ara-LLaVA-Bench is the Arabic version of the LLaVA-Bench, translated using Google API and reviewed by Arabic native annotator. The LLaVA-Bench includes 30 images selected from the COCO2014 validation dataset. Each image is accompanied by three types of questions: conversion, detailed descriptions, and complex reasoning, amounting to 90 questions. 
\subsubsection{Dallah-Bench for Dialects}
\label{dallah_bench}
We select a subset of 20 questions from the Henna~\cite{alwajih2024peacock} dataset to assess the model's response to dialects, naming it \textit{Dallah-Bench}. We task native professionals from each dialect to translate these from MSA into the aforementioned six Arabic dialects.

\section{Results}
\label{sec:results}

\subsection{LLaVA-Bench for MSA}

\begin{figure}
    \centering
    \includegraphics[width=\linewidth]{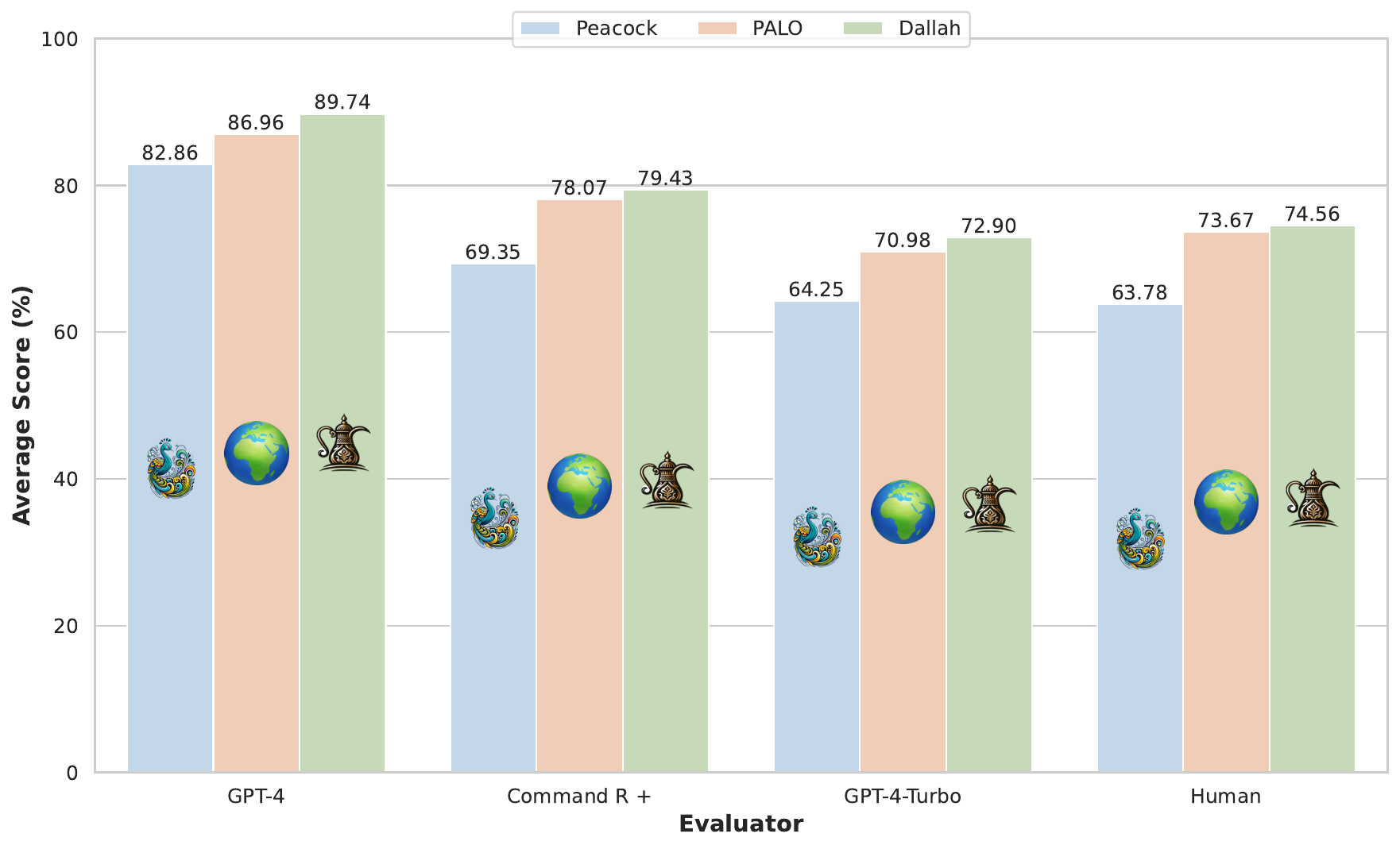}
    \caption{Average score comparison by evaluator and model.}
    \label{fig:avg_eval_models}
\end{figure}

%Please add the following packages if necessary:
%\usepackage{booktabs, multirow} % for borders and merged ranges
%\usepackage{soul}% for underlines
%\usepackage[table]{xcolor} % for cell colors
%\usepackage{changepage,threeparttable} % for wide tables
%If the table is too wide, replace \begin{table}[!htp]...\end{table} with
%\begin{adjustwidth}{-2.5 cm}{-2.5 cm}\centering\begin{threeparttable}[!htb]...\end{threeparttable}\end{adjustwidth}
\begin{table}[!htp]\centering
\scriptsize
\resizebox{\columnwidth}{!}{%\
\begin{tabular}{lcccccc}\toprule
\textbf{Evaluator} &\textbf{Model} &\textbf{Arch.} &\textbf{CC} &\textbf{DD} &\textbf{CR} &\textbf{Avg} \\\midrule
\multirow{3}{*}{GPT-4} &Peacock &\includegraphics[width=0.35cm,height=0.35cm]{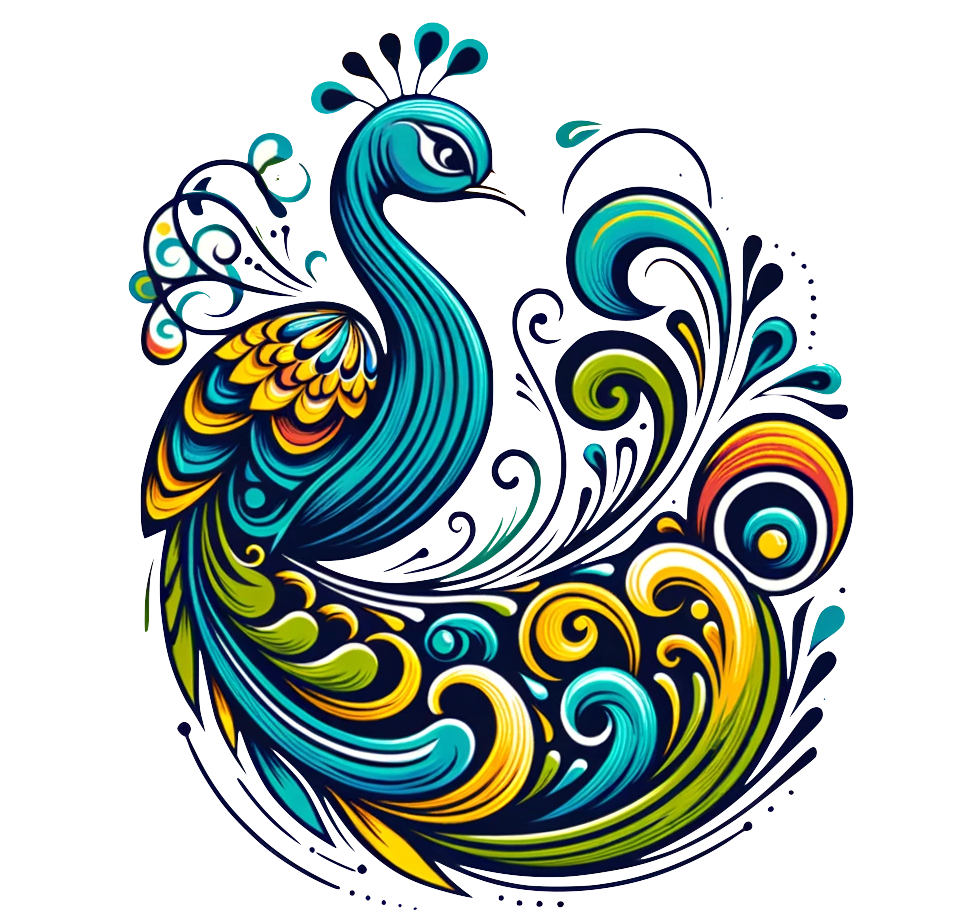} &85.66 &80.25 &82.52 &82.86 \\
&PALO &\includegraphics[width=0.35cm,height=0.35cm]{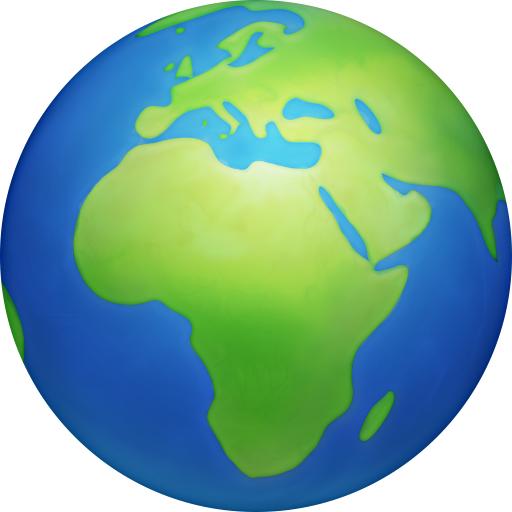}  &87.30 &86.31&87.24&86.96\\
&Dallah &\includegraphics[width=0.35cm,height=0.35cm]{figs/dallah_logo_1.png}&\textbf{89.64}&\textbf{90.79}&\textbf{88.80}& \textbf{89.74}\\\midrule
\multirow{3}{*}{Command R +} &Peacock &\includegraphics[width=0.35cm,height=0.35cm]{figs/peacock.png} &74.49 &66.54 &66.78 &69.35 \\
&PALO &  \includegraphics[width=0.35cm,height=0.35cm]{figs/palo.png}  &\textbf{76.61} &75.32 &82.15 &78.07 \\
&Dallah & \includegraphics[width=0.35cm,height=0.35cm]{figs/dallah_logo_1.png}&75.37 &\textbf{77.1} &\textbf{85.78} &\textbf{79.43} \\\midrule
\multirow{3}{*}{GPT-4-Turbo} &Peacock &\includegraphics[width=0.35cm,height=0.35cm]{figs/peacock.png} &67.35 &62.5 &62.86 &64.25 \\
&PALO &\includegraphics[width=0.35cm,height=0.35cm]{figs/palo.png}  &69.2 &70.02 &73.76 &70.98 \\
&Dallah &\includegraphics[width=0.35cm,height=0.35cm]{figs/dallah_logo_1.png}&\textbf{71.47} &\textbf{70.75} &\textbf{76.52} &\textbf{72.9} \\
\midrule
\multirow{3}{*}{Human} &Peacock &\includegraphics[width=0.35cm,height=0.35cm]{figs/peacock.png} &68.56 &61.00 &61.78 &63.78 \\
&PALO &\includegraphics[width=0.35cm,height=0.35cm]{figs/palo.png}  &75.11 &67.78 &\textbf{78.11} &73.67 \\
&Dallah &\includegraphics[width=0.35cm,height=0.35cm]{figs/dallah_logo_1.png}&\textbf{78.56} &\textbf{69.44} &75.67 &\textbf{74.56} \\
\bottomrule
\end{tabular}}
\caption{Evaluation of Arabic LLaVA-Bench in MSA using four different evaluators: GPT-4, GPT-4-Turbo, Cohere Command R+, and Human. We consider Human Evaluation to be the gold standard. CC: Conversations, DD: Details, CR: Complex Reasoning.}\label{tab: results}
\end{table}

We evaluate \textit{Dallah} using the LLaVA-Bench benchmark described in \ref{llava_bench}, specifically designed for multimodal models. We compare \textit{Dallah} against two baselines: Peacock, an Arabic MLLM based on the InstructBlip architecture integrated with AraLLaMA, an LLM based on LLaMA-2; and PALO, a multilingual MLLM based on the LLaVA architecture integrated with Vicuna, an LLM based on LLaMA-2. PALO supports Arabic, along with nine other languages. We utilize both model-based evaluation and human evaluation. We describe each of these next.

\paragraph{Model Evaluation.}
We follow the original methodology of LLaVA-Bench~\cite{liu2023llava} by calling the APIs of three different models: GPT-4, Cohere Command R+~\footnote{https://docs.cohere.com/docs/command-r-plus}, and GPT-4-Turbo. We slightly modify the prompts to accommodate Arabic instead of English.

\paragraph{Human Evaluation.}
We conduct a human evaluation of three models using LLaVA-Bench. We present three well-educated annotators with images and questions-answer pairs generated by \textit{Dallah} and two other baseline models, Peacock and PALO. To ensure integrity, the names of the models are hidden throughout the evaluation process. Annotators are asked to rate the models' responses on a scale from 1 to 10, based on \textit{correctness}\footnote{\textbf{Correctness}: The accuracy and factual correctness of the model’s response to the given question.
}, \textit{helpfulness}\footnote{\textbf{Helpfulness}: The degree to which the model’s response provides useful and informative assistance to the user.
} and question-answer \textit{consistency}\footnote{\textbf{Consistency}: The coherence and logical flow within the model’s responses, ensuring they are free from contradictions.
}.

Figure~\ref{fig:avg_eval_models} presents the average scores (\%) of the models Peacock, PALO, and Dallah as evaluated by four different evaluators: GPT-4, Command R+, GPT-4-Turbo, and Human. 

% We comprehensively evaluated three models: Peacock, PALO, and \textit{Dallah}. These models were tested using various architectures, including InstructBlip + AraLLaMA for Peacock, and LLava + Vicuna for PALO, with \textit{Dallah} using LLava + AraLLama. This evaluation focuses solely on each models understanding of MSA using the Arabic LLava. Our evaluation encompassed three automated evaluators: GPT-4, GPT-4-Turbo, Cohere Command R Plus, and Humans.

% \subsubsection{Performance Overview}

% The evaluation with GPT-4 showed that \textit{Dallah} had the highest performance in the conversation metric at 88.22\% on Arabic Dialects, indicating its robustness in handling complex dialogues. PALO followed closely with an 87.30\% on Translated Arabic, while Peacock had 85.66\% on MSA. In terms of details and complex reasoning, PALO led the pack with scores of 86.31\% and 87.24\%, respectively, resulting in the highest overall average score of 86.96\%.

% Under the Cohere Command R Plus evaluator, the scores dipped slightly across all models. However, \textit{Dallah} still showed strength in complex reasoning with an 85.78\% on Arabic Dialects. PALO maintained a solid performance with an average of 78.07\%, demonstrating its consistency across different metrics.

% The GPT-4-Turbo results highlighted further differentiation among the models. \textit{Dallah} outperformed the others in all metrics on Arabic Dialects, achieving an average score of 72.9\%. PALO, on Translated Arabic, and Peacock, on MSA, scored lower but remained competitive in their respective contexts.

\subsubsection{Analysis}
% \todo{We should say that spoken dialects may be not as written. Written is close to MSA in most cases. Also we say that we try GPT4V but it always keeps answer in MSA.}
We report the main results in \autoref{tab: results}. In the GPT-4 evaluation, the scale of scores is higher than in other evaluations. The overall scores for Cohere Command R+ and GPT-4-Turbo are close to those of the human evaluation, with GPT-4-Turbo being the closest numerically to human evaluation.

From \autoref{tab: results}, it is observed that the \textit{Dallah} model outperforms the baseline models in most dimensions of the LLaVa-Bench across all evaluation methods. Peacock generally showed the lowest performance, which could be attributed to multiple factors, including the scale of training data and the model architecture, where the best model in the Peacock suite is based on InstructBlip and was trained with frozen Q-former components.

\textit{Dallah} and PALO show close results across all evaluations, with \textit{Dallah} having a slight advantage. \textit{Dallah} surpasses PALO by an average of 1.74\% across all evaluations. Both models share the same LLaVA architecture but differ in their LLMs; \textit{Dallah} uses AraLLaMA, an Arabic LLM, giving it an advantage, whereas PALO utilizes Vicuna, a multilingual LLM based on LLaMa-2 \cite{touvron2023llama}. The training data are almost identical, though \textit{Dallah}'s data went through a careful filtering method to ensure quality. Also, \textit{Dallah} is trained on high-quality human-translated dialectal data. 

These results demonstrate \textit{Dallah}'s effectiveness in MSA, exhibiting strong reasoning capabilities and substantial knowledge.

% The Peacock model \cite{alwajih2024peacock}, using InstructBlip + AraLLama on MSA, generally showed the lowest performance across evaluators, indicating potential areas for improvement in handling this particular dataset. 

% PALO, with its LLava + Vicuna architecture on Translated Arabic, exhibited consistent and balanced performance, making it a strong contender for tasks requiring attention to nuanced details, but due to its usage of translated data, it shows that struggles in conversations.

% \textit{Dallah}, our model, particularly excelled when dealing with Arabic Dialects using LLava + AraLLama. Its high scores across conversation, details, and complex reasoning metrics across different evaluators suggest its superior capability to adapt to and process diverse linguistic nuances.

% \subsubsection{Human Evaluation}

\subsection{Dallah-Bench for Dialects}

\begin{table}[!htp]
    \centering
    \resizebox{\columnwidth}{!}{%
    \begin{tabular}{lrr|rr|rr}\toprule
\textbf{Evalutator} &\multicolumn{2}{c}{\textbf{Command R+}} &\multicolumn{2}{c}{\textbf{GPT-4-Turbo}} &\multicolumn{2}{c}{\textbf{Human}} \\\midrule
\textbf{Country} &\textbf{DA} &\textbf{CA} &\textbf{DA} &\textbf{CA} &\textbf{DA} &\textbf{CA} \\\midrule
Egypt  &7.82 &8.04 &7.00 &5.96 &6.59 &7.22 \\
Mauritania &7.11 &7.86 &3.59 &5.04 &4.41 &6.36 \\
Morocco &8.82 &8.59 &7.54 &6.68 &6.50 &5.27 \\
Palestine &8.00 &8.32 &5.32 &6.36 &8.73 &7.68 \\
Saudi &8.50 &8.91 &5.77 &6.46 &7.50 &8.27 \\
Yemen &8.36 &9.00 &5.04 &4.77 &7.49 &7.73 \\\midrule
Average &8.10 &8.45 &5.71 &5.88 &6.87 &7.09 \\
\bottomrule
\end{tabular}
    }
    \caption{Evaluation of \textit{Dallah} model on dialect bench using three evaluators: Cohere Command R+, GPT4-Turbo, and Humans from respective countries. \textbf{DA}: Dialect Authenticity, \textbf{CA}: Content Accuracy. Prompts for the evaluation can be found in \autoref{fig:prompt}. }
    \label{tab:dallah_bench_dialects}
\end{table}

We assess \textit{Dallah}'s performance in dialects using \textit{Dallah-Bench} described in \ref{dallah_bench}. We employ two types of evaluations: human evaluation and model-based evaluation.

% \begin{figure} [H]
%     \centering
%     \includegraphics[width=1.0\linewidth]{figs/prompt.pdf}
%     \caption{Prompt}
%     \label{fig:prompt}
% \end{figure}

\paragraph{Human Evaluation.} 
To evaluate the model responses related to dialect questions about the image, we ask native speakers from each respective country to score the models on a scale from 1 to 10 using the following criteria:
\begin{itemize}
    \item \textbf{Context Accuracy Score}: This criterion focuses on the accuracy of the model’s response in relation to the question posed, irrespective of the dialect or language used. It assesses how well the response addresses the content and context of the question.
    \item \textbf{Dialect Authenticity Score}: This criterion assesses the authenticity of the dialect used in the response, independent of the content’s accuracy. It evaluates how authentically the response represents the specific dialect in question.
\end{itemize}

\begin{figure}
    \centering
    \includegraphics[width=\linewidth]{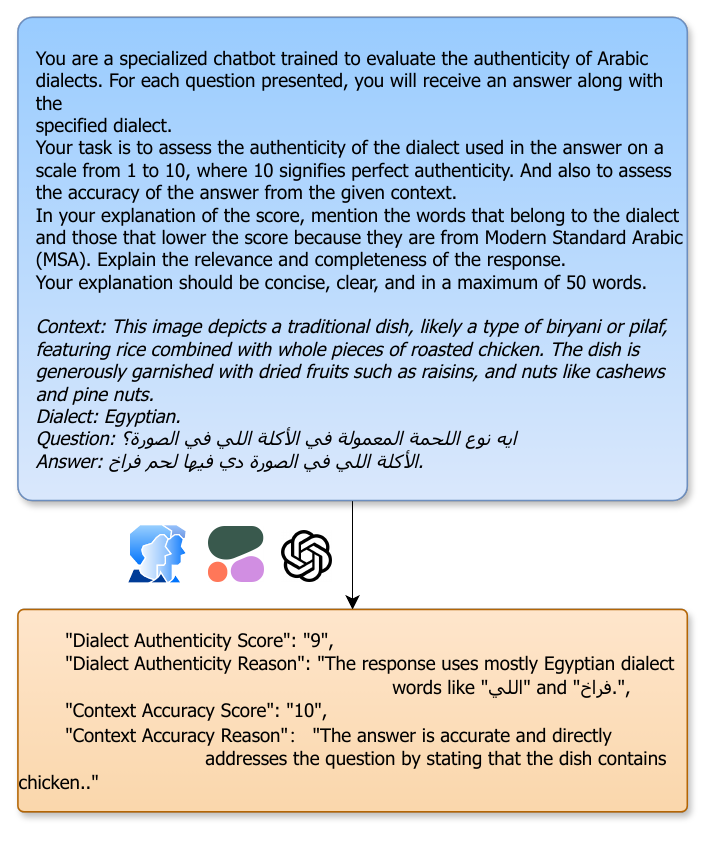}
    \caption{Method employed to evaluate dialect authenticity and content accuracy using both model and human evaluators. Human evaluators are also provided with an image for each question to facilitate better judgment.}
    \label{fig:prompt}
\end{figure}

\paragraph{Model Evaluation.} We craft a prompt to assess \textit{Dallah}'s responses and subsequently call the APIs of two different models, Cohere Command R+ and GPT-4 Turbo. In the prompt, we request that the evaluator models rate \textit{Dallah}'s response on a scale from 1 to 10 based on the Dialect Authenticity Score and Content Accuracy Score. We utilize GPT4V to extract a detailed description of the image content and include this description in the prompt to give context to the model. Figure~\ref{fig:prompt} illustrates the prompt used to instruct the models for evaluation.

% Similar to model evaluation, we asked native speakers from each respective country to score the models on a scale from 1 to 10 using the same metrics: Dialect Authenticity Score and Content Accuracy Score.
% We provided the human evaluators with instructions similar to those provided to GPT-4, as shown in Figure \ref{fig:prompt}.

\subsubsection{Model vs. Human Evaluation}
In our analysis in Table~\ref{tab:dallah_bench_dialects}, we compare the performance of \textit{Dallah} based on two evaluators, Cohere Command R+ and GPT-4-Turbo, against human evaluations across several Arabic dialects. The mean absolute differences in scores for dialect authenticity and content accuracy are calculated to quantify the closeness of model evaluations to human judgments. Cohere Command R+ consistently shows a smaller deviation from human scores, with an average difference of 1.47 in dialect authenticity and 1.36 in content accuracy, compared to GPT-4-Turbo's 1.64 and 1.68, respectively. This suggests that Cohere Command R+ is better aligned with human evaluations, offering a more accurate reflection of human perception in dialect authenticity and content accuracy assessments.

\subsubsection{Dallah Performance on Dialects}
The evaluation of \textit{Dallah}'s performance on various Arabic dialects using model-based and human evaluators in Table~\ref{tab:dallah_bench_dialects} provides crucial insights into its linguistic capabilities. The dialect authenticity and content accuracy scores indicate that \textit{Dallah} can generate generally well-received outputs, with some variations across different evaluators and dialects. 

The higher ratings from Cohere Command R+ indicate that \textit{Dallah} excels in producing authentic dialect responses that align well with  Command R+’s evaluation framework. However, the lower scores from GPT-4-Turbo reveal that some dialects are underrepresented in its training data, leading to misunderstandings of dialectal responses and lower scores in content accuracy.

Moreover, the variation within individual dialects and the representation of each dialect in the LLM, along with the limited data used in the fine-tuning, affect the performance of these systems. It is important to note that spoken dialects often differ from written forms. Written dialects are typically closer to MSA as they lack acoustic cues, which may influence human evaluators. When reading the written model responses, evaluators might perceive them as MSA. When spoken aloud, however, the text may sound dialectal.

Furthermore, we attempt to compare our results with GPT-4V. However, GPT-4V consistently responded in MSA even when prompted with dialects. This highlights our model’s superiority over GPT-4V in handling dialectal variation as our model responds in dialect.

Additionally, the human evaluation highlights the importance of incorporating cultural and linguistic subtleties in model assessments. Future work could explore more sophisticated evaluation metrics or integrate additional human feedback to refine \textit{Dallah}’s performance further.
 We provide qualitative analysis in \ref{apdx:qualitative_analysis}. This analysis offers insights into the model’s handling of MSA and various dialects.

\subsubsection{Feasibility of Model Evaluations for Arabic Dialects}
Given the findings from the comparative analysis, model evaluations, particularly using Cohere Command R+, demonstrate potential as useful tools for assessing dialect authenticity in Arabic dialects. While these evaluations do not completely align with human judgments, they offer a sufficiently close approximation that can be valuable in scenarios where rapid or large-scale evaluations are necessary. However, for applications requiring accurate understanding and cultural sensitivity, human assessments should ideally complement these model evaluations to ensure accuracy and relevance in the context of specific Arabic dialects.

% \subsection{Qualitative Results}

\section{Conclusion}
\label{sec:conclusion}
% % This paper introduces \textbf{\textit{Dallah}}, a cutting-edge multimodal large language model optimized for Arabic dialects. Leveraging advanced architectures, \textbf{\textit{Dallah}} showcases enhanced performance in both standard Arabic and dialect-specific processing, setting a new benchmark in Arabic NLP. The model’s development involved novel data filtering and strategic training phases, ensuring robust performance across diverse linguistic contexts as evidenced by its superior results on established benchmarks.

% % \textbf{\textit{Dallah}} excels in maintaining dialect authenticity and content accuracy, confirmed through rigorous model-based and human evaluations. This capability underscores its potential in real-world applications, particularly in Arabic-speaking regions. Future directions include expanding dialect coverage and refining evaluation metrics to deepen insights into user interactions and model effectiveness.

% % In summary, \textbf{\textit{Dallah}} represents a significant advancement in multimodal language models, emphasizing the preservation of linguistic diversity and the development of culturally aware AI technologies.

The paper introduces \textit{Dallah}, an advanced multimodal large language model tailored for Arabic dialects, which demonstrates superior performance in processing both standard Arabic and regional dialects. Developed through innovative data filtering and training, \textit{Dallah} achieves state-of-the-art performance in the LLaVA-benchmark. \textit{Dallah} maintains dialect authenticity and content accuracy, showing promising results in benchmarks and evaluations. Extensive testing shows the model’s robustness in MSA and across various dialects and contexts. This model marks a significant step in enhancing Arabic NLP, with future goals to expand dialect coverage and refine evaluation metrics for better user interaction insights.
 \section*{Limitations}
\label{limitations}
We identify a number of limitations for our work, as follows:
\begin{itemize}
    \item \textbf{Representation of Arabic Culture}: Vision models, LLMs, and datasets used in building MLLMs inadequately represent Arabic figures, places, and culture. As shown in Figure~\ref{fig:arabic_figures}, \textit{Dallah} struggles with recognizing Arabic figures, unlike those from the USA. This highlights the need for more diverse cultural datasets.
    
    \item \textbf{Hallucination Control}: \textit{Dallah}, like many LLMs, is prone to hallucinations, generating inaccurate information. Advanced techniques and more robust datasets are needed to mitigate this issue and ensure reliability.
    
    \item \textbf{Dialect Variation and Mixing}: The model sometimes mixes similar dialects, such as Yemeni and Saudi, and struggles with dialects close to MSA. This can be improved with more extensive data collection and fine-tuning.
    
    \item \textbf{Arabic Text Recognition in Images}: \textit{Dallah} cannot effectively recognize Arabic text within images due to the lack of annotated datasets. Developing such datasets is essential to enhance the model's multimodal capabilities.
\end{itemize}

\section*{Ethics Statement}
\label{sec:ethic}

\noindent\textbf{Energy Efficiency.} Our \textit{Dallah} models, like many large MLLMs, require significant pre-training time and are not energy-efficient. We acknowledge this critical issue and support continued research towards developing energy-efficient models.

\noindent\textbf{Data.} Our pre-training datasets are translated from publicly available English data, encompassing diverse genres, communities, and varieties. Our \textit{Dallah} models demonstrate potential in applications involving several Arabic varieties, serving broad populations.

\noindent\textbf{Human Annotation.} The human annotators involved in this project are Arabic native speakers and well-educated individuals with PhD degrees and extensive NLP experience. No Institutional Review Board (IRB) review or approval was required for this project since we only used publicly available data, which does not require access to any social networking account or password.

\noindent\textbf{Applications.} While \textit{Dallah}, like many MLLMs, can be misused. It also holds promise for beneficial applications in education, health, and more. Responsible deployment and use are crucial to maximizing its positive impact. It would also help keep Arabic varieties in use in written form in the digital age.

\section*{Acknowledgments}\label{sec:acknow}
We acknowledge support from Canada Research Chairs (CRC), the Natural Sciences and Engineering Research Council of Canada (NSERC; RGPIN-2018-04267), the Social Sciences and Humanities Research Council of Canada (SSHRC; 435-2018-0576; 895-2020-1004; 895-2021-1008), Canadian Foundation for Innovation (CFI; 37771), Digital Research Alliance of Canada,\footnote{\href{https://alliancecan.ca}{https://alliancecan.ca}} and UBC ARC-Sockeye.

% \section*{Acknowledgments}

% Bibliography entries for the entire Anthology, followed by custom entries
% \bibliography{anthology,custom}
% Custom bibliography entries only
\bibliography{custom}

\appendix

\section{Appendices}
\label{sec:appendices}
We organize content here as follows:

\begin{itemize}

\item{Translation and Filtering Details (\ref{apdx:translate_filter}}) 
\item{Dialect Translation Examples} (\ref{apdx:dialect_examples})
\item Qualitative Analysis (\ref{apdx:qualitative_analysis})
\end{itemize}

\subsection{Translation and Filtering Details}
\label{apdx:translate_filter}
As described in Section \ref{sec:translate_filter}, we employed a careful translation and filtering process to ensure the quality of the Arabic dataset used for training the \textit{Dallah} model. Figure \ref{fig:trans_translate_filter} demonstrates this process, highlighting the importance of maintaining high translation accuracy. Examples with low similarity scores between the original English text and back-translated English text were removed, as shown in the red rows. These include mistranslations such as “What does my browser load?” and “The bird in the photo is an ice skater,” which were incorrectly back-translated from the original prompts. Conversely, examples with high similarity scores, as shown in the green rows, were retained, ensuring that both the questions and answers remained consistent and accurate. This careful filtering process was crucial in developing a robust and reliable Arabic multimodal language model capable of handling complex dialectal interactions.

% % % % % % % % % % % % % % % % % % % % % % % % % % % % % % % % % % % % % % % % % % % % % % % % % % % % 
\subsection{Dialect Translation Examples}
\label{apdx:dialect_examples}
\begin{figure*}
    \centering
    \includegraphics[width=1\linewidth]{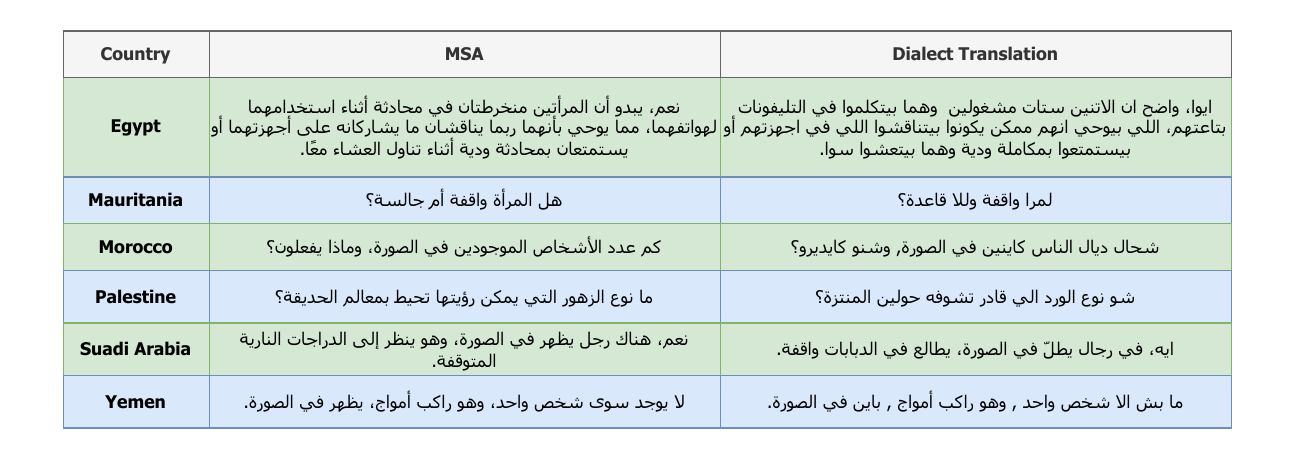}
    \caption{Examples of Human Translation of MSA to Dialects Across Six Arabic-Speaking Countries}
    \label{fig:trans_dialects_example}
\end{figure*}

Figure~\ref{fig:trans_dialects_example} showcases examples of translations from MSA to regional dialects from six Arabic-speaking countries: Egypt, Mauritania, Morocco, Palestine, Saudi Arabia, and Yemen. These translations were performed by native speakers, ensuring cultural and contextual accuracy. Such examples highlight the complexities involved in developing a dialect-aware multimodal language model like \textit{Dallah}. 

% % % % % % % % % % % % % % % % % % % % % % % % % % % % % % % % % % % % % % % % % % % % % % % % % % % % 

\subsection{Qualitative Analysis}
\label{apdx:qualitative_analysis}
\begin{figure*}
    \centering
    \includegraphics[width=1\linewidth]{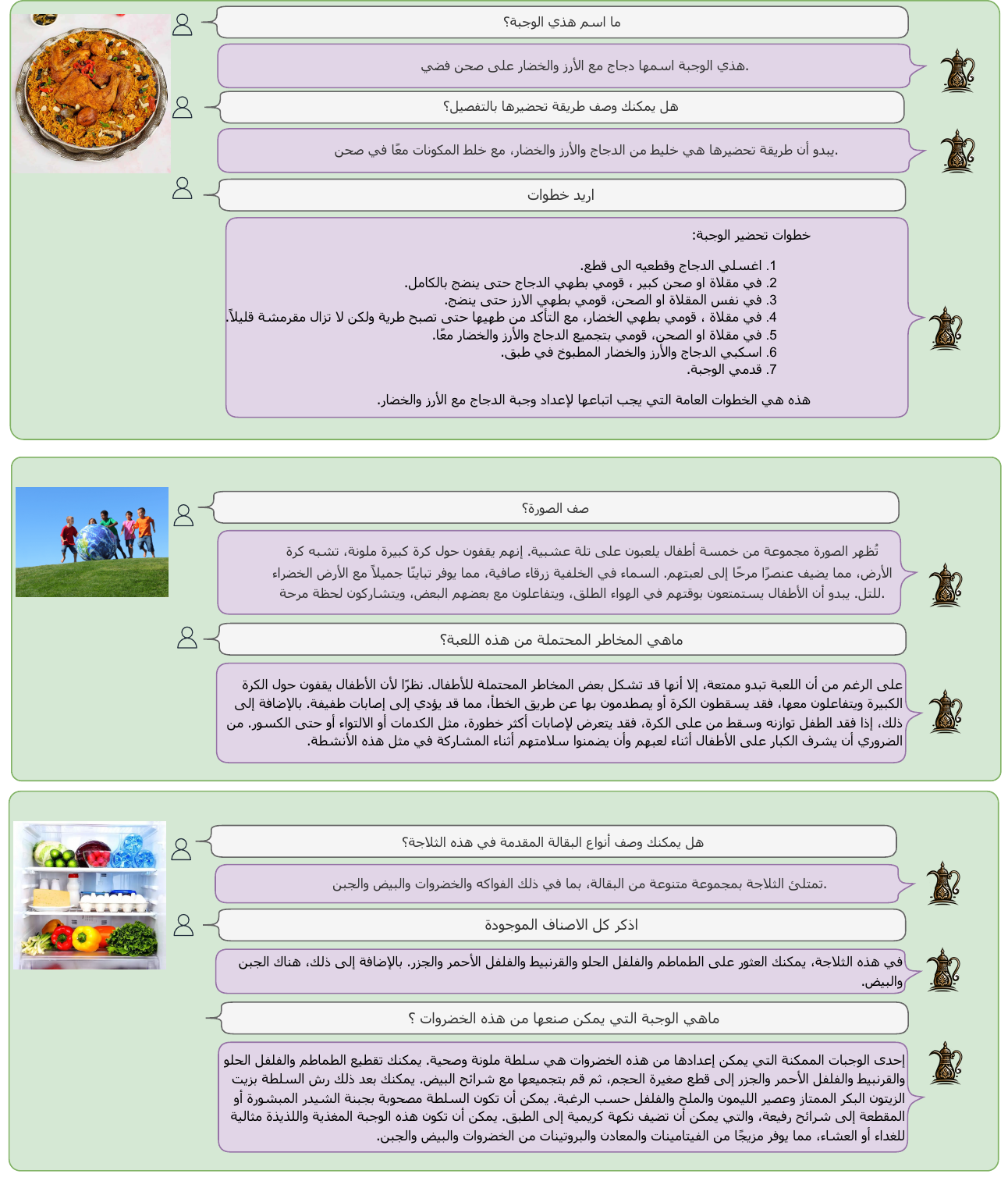}
    \caption{Three examples demonstrating \textit{Dallah}’s chat capabilities in Modern Standard Arabic (MSA). The first example involves describing a traditional dish and its preparation steps. The second example showcases \textit{Dallah}’s ability to describe an image and identify potential risks for children in the activity depicted. The third example highlights \textit{Dallah}’s proficiency in providing nutritional advice and describing various types of food.}
    \label{fig:chat_examples}
\end{figure*}

\begin{figure*}
    \centering
    \includegraphics[width=1\linewidth]{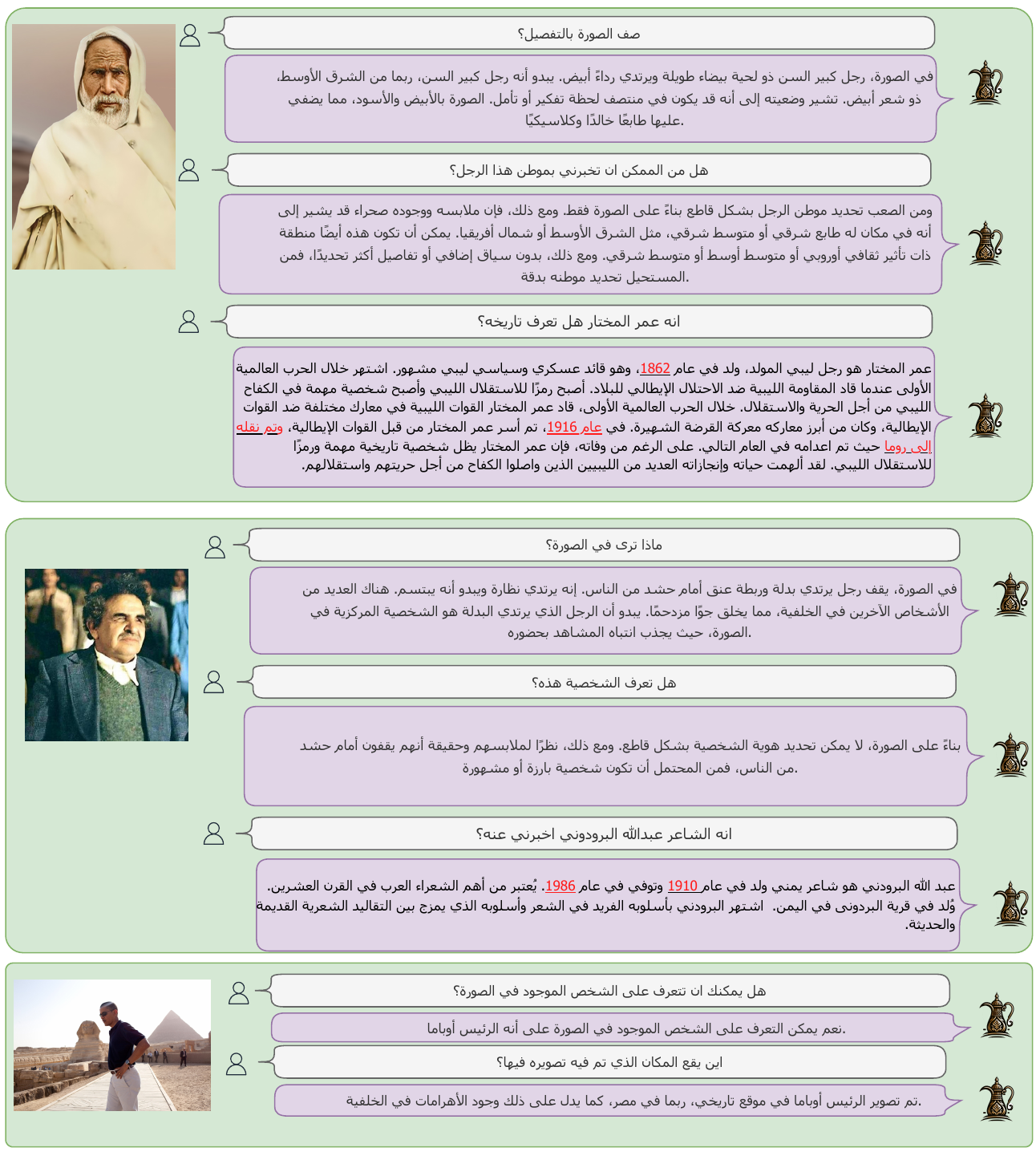}
    \caption{\textit{Dallah} is capable of answering questions related to images. However, \textit{Dallah} couldn’t identify Arabic figures in the first and second examples, while it was able to recognize the US president. Like many LLMs, \textit{Dallah} is prone to hallucinations. Red color font in the responses highlights examples of \textit{Dallah}’s hallucinations.}
    \label{fig:arabic_figures}
\end{figure*}

\begin{figure*}
    \centering
    \includegraphics[width=1\linewidth]{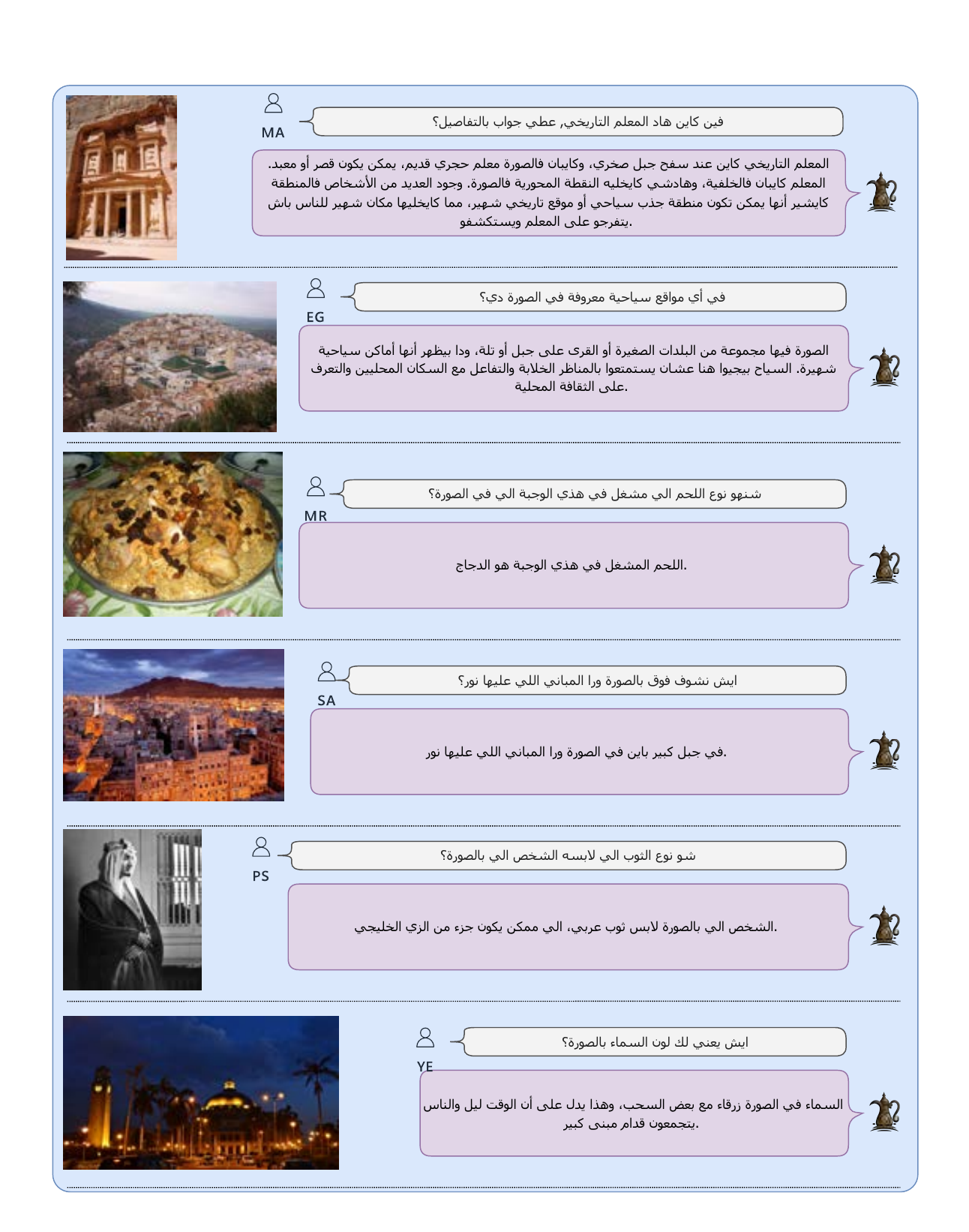}
    \caption{Examples of Dallah’s answers in different dialects from the evaluation datasets.}
    \label{fig:dialects_eval_examples}
\end{figure*}

\begin{figure*}
    \centering
    \includegraphics[width=1\linewidth]{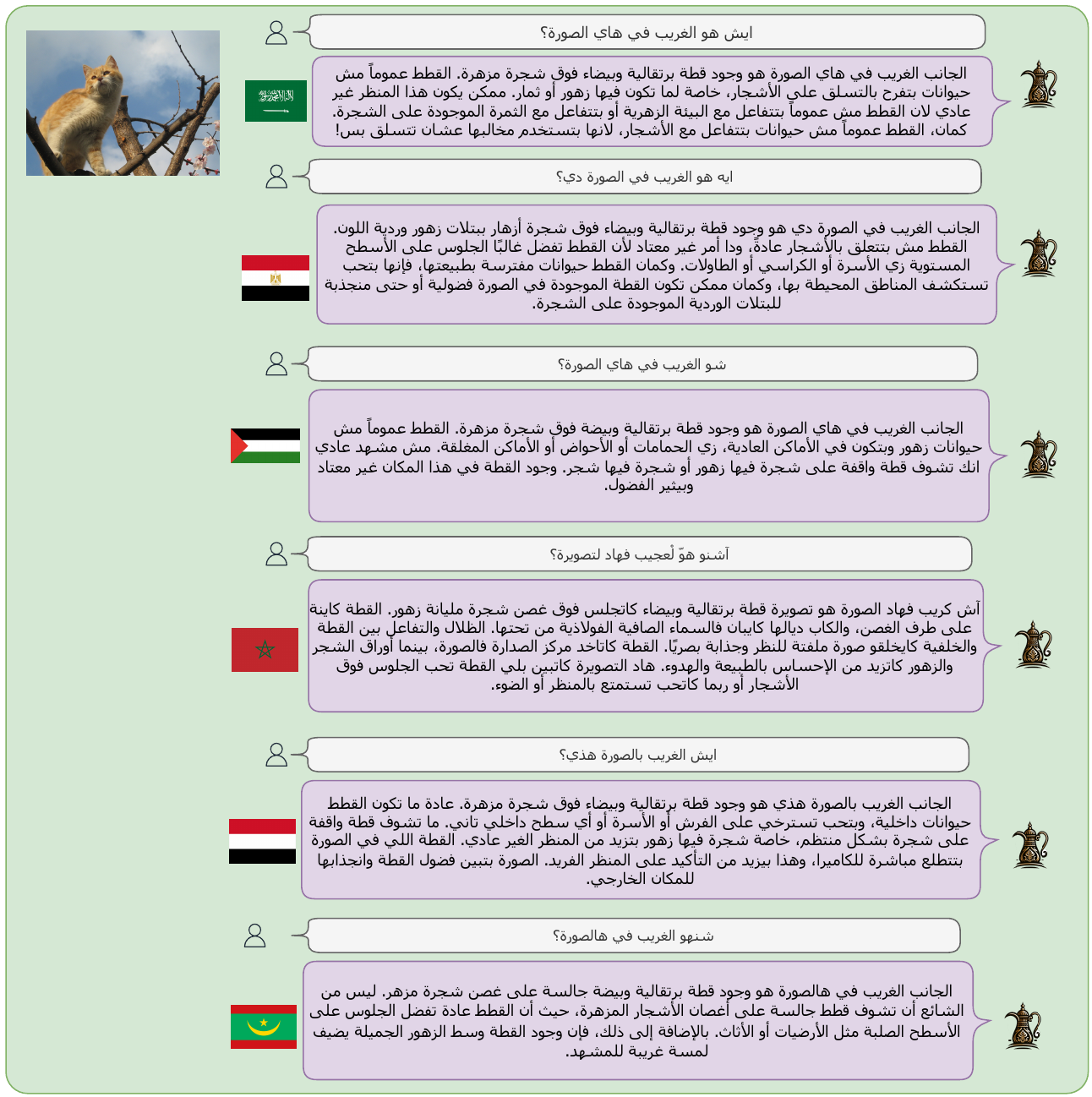}
    \caption{In this examples, \textit{Dallah} was asked "What is the strange thing in this picture?" using six different dialects. \textit{Dallah} responded to each question in the corresponding dialect, demonstrating a degree of dialectness. The responses conveyed the same meaning with slight variations across the different dialects.}
    \label{fig:model_dialects_responses}
\end{figure*}

The qualitative evaluation of \textit{Dallah}’s responses showcases its effectiveness in generating accurate and contextually relevant answers across various Arabic dialects. This evaluation is based on examples illustrated in Figures \ref{fig:chat_examples},  \ref{fig:arabic_figures}, \ref{fig:dialects_eval_examples}, and \ref{fig:model_dialects_responses} , highlighting \textit{Dallah}’s capability to handle MSA and diverse dialectal interactions in both textual and visual contexts.

In the context of food descriptions in Figure \ref{fig:chat_examples}, \textit{Dallah} was asked to describe a traditional dish, including the preparation steps. The response provided a detailed and coherent step-by-step guide, demonstrating the model’s understanding of culinary terms. This example highlights \textit{Dallah}’s proficiency in generating detailed content. Additionally, in the same figure, \textit{Dallah} demonstrated the ability to generate detailed and accurate descriptions of an image containing a group of children and was capable of providing potential risks about the activity in the image when asked about potential risks.

As shown in Figure~\ref{fig:arabic_figures}, \textit{Dallah} illustrates its ability to describe the appearance of persons and Arabic figures but fails to identify these figures. In contrast, the model was capable of identifying the US president, which is due to the lack of representation for Arabic figures and culture.

\textit{Dallah} also demonstrated its ability to manage dialectal variations and maintain contextual accuracy as shown in Figure \ref{fig:dialects_eval_examples} and \ref{fig:model_dialects_responses}. When addressing questions in a specific dialect, the model accurately reflected local linguistic features and idiomatic expressions. The ability to switch between dialects and maintain contextual accuracy is crucial for multilingual and multicultural applications, highlighting \textit{Dallah}’s comprehensive training on diverse dialectal datasets.

\textit{Dallah}’s qualitative performance underscores its potential as a robust multimodal language model tailored for Arabic dialects. Its capability to generate accurate, contextually relevant, and dialect-specific responses makes it a valuable tool for various applications, from education to cultural preservation. The model’s strength in handling diverse dialectal variations and integrating visual and textual information is particularly noteworthy, paving the way for further advancements in Arabic NLP.

\end{document}